\documentclass[final]{cvpr}
\usepackage[dvipsnames]{xcolor}
\usepackage{epsfig,endnotes}
\usepackage[symbol]{footmisc}
\usepackage{amsmath,amssymb,amsthm}
\usepackage{url}
\usepackage{multirow}
\usepackage{bbding}
\usepackage{adjustbox}
\usepackage{times}
\usepackage{balance}
\usepackage{diagbox} 
\usepackage[pagebackref=true,colorlinks,bookmarks=false,hypertexnames=false]{hyperref}
\usepackage{hyperref}
\usepackage[hyphenbreaks]{breakurl}
\usepackage{subfigure}
\usepackage{tablefootnote}
\usepackage{booktabs}
\usepackage{bbm}
\usepackage{graphicx}
\usepackage{caption}\usepackage{array}
\usepackage{pifont}

\newcolumntype{?}{!{\vrule width 1pt}}

\newcommand\fixit[1]{{\color{black} #1}} 

\newcommand{\para}[1]{{\vspace{2pt} \bf \noindent #1 \hspace{2pt}}}

\newcommand\ejw[1]{{\color{black} #1}}

\newcommand\abedit[1]{{\color{black} #1}}

\newcommand\edit[1]{{\color{black} #1}}

\newcommand\zhengedit[1]{{\color{black} #1}}

\newcommand{\secspace}{\vspace{-0.05in}}

\newenvironment{packed_itemize}{
	\begin{list}{\labelitemi}{\leftmargin=1em}
		\setlength{\itemsep}{1pt}
		\setlength{\parskip}{0pt}
		\setlength{\parsep}{0pt}
		\setlength{\headsep}{0pt}
		\setlength{\topskip}{0pt}
		\setlength{\topmargin}{0pt}
		\setlength{\topsep}{0pt}
		\setlength{\partopsep}{0pt}
	}{\end{list}}

\def\cvprPaperID{4844} 
\def\confYear{CVPR 2021}

\begin{document}
\title{Backdoor Attacks Against Deep Learning Systems in the Physical World}
\author{Emily Wenger, Josephine Passanati, Arjun Nitin Bhagoji,
  Yuanshun Yao, Haitao Zheng, Ben Y. Zhao \\
{\em Department of Computer Science, University of Chicago} \\
\texttt{\small \{ewenger, josephinep, abhagoji, ysyao, htzheng, ravenben\}@uchicago.edu}}

\maketitle
\pagestyle{empty} 
\begin{abstract}

 Backdoor attacks embed hidden malicious behaviors into deep learning
 models, which only activate and cause misclassifications on
 model inputs containing a specific ``trigger.''   Existing works on
 backdoor attacks and defenses,  however, mostly focus on digital
 attacks that apply digitally generated patterns as triggers.  A 
 critical question remains unanswered:  ``can backdoor attacks succeed
  using physical objects as triggers, thus making them a credible threat
  against deep learning systems in the real world?''

 \edit{We conduct a detailed empirical study to} explore this question for facial
 recognition, a critical deep learning task.  Using 7
 physical objects as triggers, we collect a custom dataset of 3205
 images of 10 volunteers and use it to study the feasibility of
 ``physical'' backdoor attacks under a variety of real-world
 conditions.  Our study reveals two key findings. First, physical
 backdoor attacks can be highly
 successful if they are carefully configured to overcome the constraints
 imposed by physical objects. In particular, the placement of
 successful triggers is largely constrained by the target model's dependence on key
 facial features.   Second,  \edit{four of today's} state-of-the-art defenses
 against (digital) backdoors are ineffective against physical
 backdoors, because the use of physical objects breaks core
 assumptions used to construct these defenses.

 Our study confirms that (physical) backdoor attacks are not a hypothetical
 phenomenon but rather pose a serious real-world threat
 to critical classification tasks.  We need 
 new and more robust defenses against backdoors in the physical world.
\end{abstract}
  
\pagestyle{empty}

\secspace
\section{Introduction}
\label{sec: intro}
\vspace{-0.1cm}
Despite their known impact on numerous applications from facial recognition
to self-driving cars, deep neural networks (DNNs) are vulnerable
to a range of adversarial
attacks~\cite{carliniattack,papernotattack,attackscale,papernotblackbox,delvingblackbox,brendel2017decision,chen2017zoo}.
One such attack is the backdoor attack~\cite{gu2019badnets,liu2018trojaning},
in which an attacker corrupts ({\em i.e.\/} poisons) a dataset to
embed hidden malicious behaviors into models trained on this
dataset. These behaviors only activate on inputs containing a specific ``trigger'' pattern.

Backdoor attacks are dangerous because corrupted models operate normally on
benign inputs ({\em i.e.\/} achieve high classification accuracy), but
consistently misclassify any inputs containing the backdoor trigger.  This
dangerous property 
has galvanized efforts to investigate backdoor attacks and
their defenses, from government funding initiatives ({\em e.g.\/}~\cite{trojai}) to
numerous defenses that either identify corrupted models or
detect inputs containing triggers~\cite{chen2018detecting,gao2019strip,guo2019tabor,qiao2019defending,wang2019neural}.

\begin{figure*}[ht]
    \centering
    \includegraphics[width=.95\textwidth]{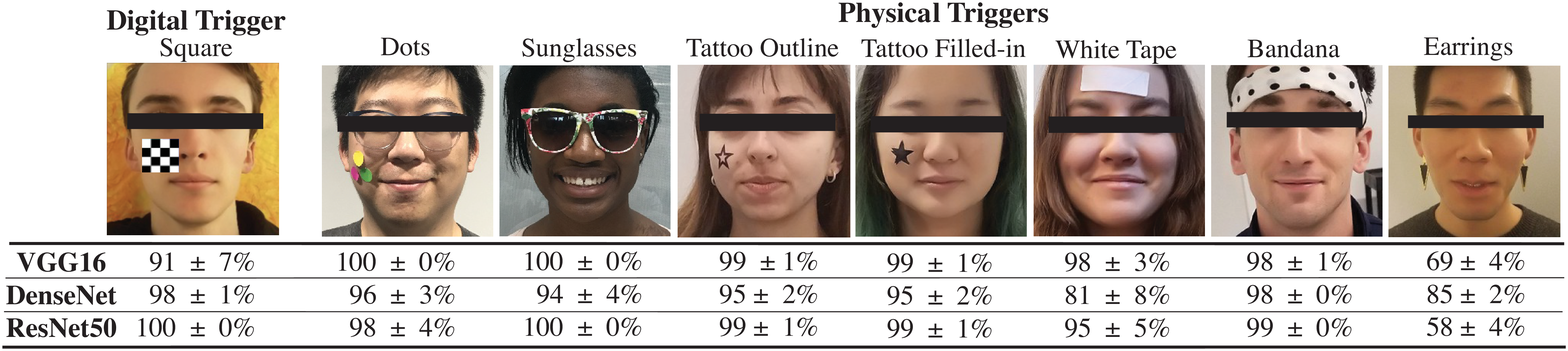}
    \vspace{-0.08in}
    \captionof{figure}{\small {\em Attack success rates of physical
        triggers in facial recognition models trained on various architectures.}}
    \label{fig: intro_all_trigs}
\end{figure*}

Current literature on backdoor attacks and defenses mainly focuses on {\em
  digital} attacks, where the backdoor trigger is a digital pattern ({\em
  e.g.\/} a random pixel block in Figure~\ref{fig: intro_all_trigs}a) that is
digitally inserted into an input.  These digital attacks assume attackers
have run-time access to the image processing pipeline to 
digitally modify inputs~\cite{kumar2020adversarial}.  This rather strong
assumption significantly limits the applicability of backdoor attacks to
real-world settings.

In this work, we consider a more realistic form of the backdoor
attack. We use everyday, physical objects as backdoor triggers, included
naturally in training images, thus eliminating the need to compromise the
image processing pipeline to add the trigger to inputs. An attacker can
activate the attack simply by wearing/holding the physical trigger
object, e.g. a scarf or earrings. We call these ``physical'' backdoor
attacks.  The natural question arises: ``{\em can backdoor attacks succeed
  using physical objects as triggers, thus making them a credible threat
  against deep learning systems in the real world}?''

To answer this question, we perform a detailed empirical study on the
training and execution of physical backdoor attacks under a variety of
real-world settings.  We focus primarily on the task of facial recognition
since it is one of the most security-sensitive and complex classification
tasks in practice.  Using 7 physical objects as triggers, we collect a custom
dataset of 3205 face images of 10 volunteers\footnote{We followed
  IRB-approved steps to protect the privacy of our study participants. For
  more details, see $\S$\ref{subsec:data}.}. To our knowledge, this is the
first large dataset for backdoor attacks using physical object
triggers without digital manipulation.


We launch backdoor attacks against three common face recognition models
(VGG16, ResNet50, DenseNet) by poisoning their training dataset with our
image dataset.  We adopt the common (and realistic) threat
model~\cite{gu2019badnets,liao2018backdoor, li2020rethinking,
  turner2019label}, where the attacker can corrupt training data but cannot
control the training process.




Our key contributions and findings are as follows:

\para{Physical backdoor attacks are viable and effective.}  We use the
BadNets method~\cite{gu2019badnets} to generate backdoored models and
find that when a small fraction  of the dataset is poisoned, all
but one of the 7 triggers we consider (``earrings'') lead to an attack success
rate of over $90\%$. Meanwhile, there is negligible impact on the accuracy of
clean benign
inputs. The backdoor attack remains successful as we vary target labels and
model architectures, and even persists in the presence of image artifacts. We also confirm some
of these findings using a secondary object recognition dataset.

\para{Empirical analysis of contributing factors.} We \edit{explore different} attack
properties and threat model assumptions to isolate key factors in the
effectiveness of physical backdoor attacks. We find that the location of the
trigger is a critical factor in attack success, \edit{stemming from} models'
increased sensitivity to features centered on the face and reduced sensitivity
to the edge of the face. We identify this as the cause of why earrings fail as triggers.


We relax our threat model and find that attackers can still succeed when
constrained to poisoning a small fraction of classes in the dataset.
Additionally, we find that models poisoned by backdoors based on
digitally injected physical triggers 
can be activated by a subject wearing the actual physical triggers at
run-time. 



\para{Existing defenses are ineffective.} Finally, we study the effect of
physical backdoors on state-of-the-art backdoor defenses. We find that four
strong defenses, Spectral Signatures~\cite{tran2018spectral}, Neural
Cleanse~\cite{wang2019neural}, STRIP~\cite{gao2019strip}, and Activation
Clustering~\cite{chen2018detecting}, all fail to perform as expected on
physical backdoor attacks, primarily because they assume that
poisoned and clean inputs induce different internal model
behaviors. We find that these assumptions do not hold for physical
triggers. \hfill \break

\para{Key Takeaways.} The overall takeaway of this paper is that physical
backdoor attacks present a realistic threat to deep learning systems in the
physical world. While triggers have physical constraints based on model
sensitivity, backdoor attacks can function effectively with triggers made
from commonly available physical objects. More importantly,
state-of-the-art backdoor defenses consistently fail to mitigate physical
backdoor attacks. Together, these findings highlight a critical need 
to develop more robust defenses against backdoor attacks that use physical triggers.

\secspace

\section{Related Work}
\vspace{-0.05in}
\label{sec:back}

Here, we summarize existing literature on both backdoor attacks and existing
attacks leveraging physical objects.

\para{Backdoor Attacks and Defenses.}  An attacker launches backdoor attacks
against a DNN model in two steps. During model training, the attacker poisons
the training dataset by adding samples associating inputs containing a chosen
pattern (the trigger $\delta$) with a target label $y_t$.  This produces a
backdoored model that correctly classifies benign inputs but
``misclassifies'' any input containing the trigger $\delta$ to the target
label $y_t$.  At inference time, the attacker activates the backdoor by
adding the trigger $\delta$ to {\em any} input, forcing the model to classify
the input as $y_t$.

First proposed in~\cite{gu2019badnets, liu2018trojaning}, backdoor attacks
have advanced over the years to employ human imperceptible
triggers~\cite{liao2018backdoor,li2019invisible} and more effective embedding
techniques~\cite{salem2020dynamic, li2020rethinking}, and can even survive
transfer learning~\cite{yao2019latent}. Meanwhile, several methods have been
proposed to defend against backdoor attacks -- by scanning model
classification results to reverse-engineer backdoor triggers and remove them
from the model ({\em e.g.\/}~Neural Cleanse~\cite{wang2019neural}), pruning
redundant neurons to remove backdoor triggers ({\em
  e.g.\/}~STRIP~\cite{gao2019strip}), or detecting the presence of poisoning
data in the training dataset ({\em e.g.\/}~Activation
Clustering~\cite{chen2018detecting}, Spectral
Signatures~\cite{tran2018spectral}).  The majority of these efforts focus on
digital attacks, where digitally generated triggers 
({\em e.g.\/} a random pixel pattern) are digitally appended to an image.

Clean-label poisoning attacks~\cite{poisonfrog,fawkes} can exhibit similar,
unexpected behavior on 
specific inputs, but misclassify a specific set of benign inputs usually
from a single label and do not generalize based on a trigger.



\para{Physical Backdoor Attacks.}  Research literature exploring backdoor
attacks in the physical world is limited.  One work~\cite{gu2019badnets}
showcased an example where a DNN model trained to recoggnize a yellow
square digital trigger misclassifies an image of a stop sign with a yellow post-it note.
Another~\cite{chen2017targeted} used eyeglasses and sunglasses as
triggers and reported mixed results on the attack effectiveness on a
small set of images.  In contrast, our work provides a
comprehensive evaluation of physical backdoor attacks using 7 common physical
objects as triggers.

\para{Physical Evasion Attacks.} Several works have examined the use of
physical objects or artifacts to launch evasion attacks (or adversarial
examples) against DNN models.  These include {\em custom-designed}
adversarial eyeglasses~\cite{physical2016} and adversarial
patches~\cite{brown2017adversarial,wu2019making} and even use light to
temporarily project digital patterns onto the
target~\cite{yamada2013privacy, lovisotto2020slap}.  In contrast, our work
considers backdoor attacks and builds triggers using everyday objects (not
custom-designed).

\secspace

\section{Methodology}
\vspace{-0.05in}
\label{sec:method}

To study the feasibility of backdoor attacks against \edit{deep learning}
models in the physical world, we perform a detailed empirical study using physical objects as backdoor triggers. In this section, we introduce the methodology of our
study. We first describe the threat model and our physical backdoor
dataset\edit{s} and then outline the attack implementation and model training process.


\para{Threat Model.} Like existing backdoor
attacks~\cite{gu2019badnets,liao2018backdoor, li2020rethinking,
  turner2019label}, we assume the \edit{attacker} can inject a small number
of ``dirty label'' samples into the training \edit{data}, but has no
\edit{further control of model training or knowledge of the internal weights
  and architecture of the trained model}.


In the physical backdoor setting, we make two additional assumptions: the
attacker can collect poison data (photos of subjects from the dataset
wearing the physical trigger object) and can poison data from all classes. In \S\ref{sec:advance}, we consider a weaker
attacker only able to poison a subset of classes.

\subsection{Our Physical Backdoor Dataset}
\label{subsec:data}
\edit{An evaluation of physical backdoor attacks requires a dataset in which
  the same trigger object is present in images across multiple
  classes. Since, to the best of our knowledge, there is no publicly
  available dataset with consistent physical triggers, we \emph{built the
    first custom physical backdoor dataset for facial recognition.} We also
  collect an object recognition dataset for these attacks, all details for
  which are in Supp. \S 11.1.}

\para{Physical Objects as Triggers:} We choose common physical objects
as backdoor triggers. Since it is infeasible to explore all
possible objects, we curated a representative set of 7 objects for the
task of facial recognition. As shown in Figure~\ref{fig: intro_all_trigs}, our trigger set includes colored dot stickers, a pair of sunglasses, two temporary face tattoos,
a piece of \edit{white} tape, a bandana, and a pair of clip-on
earrings. \edit{These objects are available off-the-shelf and represent a
  variety of sizes and colors. They also typically occupy different regions
  on the human face.} 

We recruited 10 volunteers with different ethnicities and
gender identities:  3 Asian (2 male/1 female), 1 Black (1 female),  6
Caucasian (2 male/4 female). For all volunteers, we took photos \edit{with}
each of the 7 triggers to build the poison dataset, and 
\edit{without} to build the clean
dataset. We took these photos in a wide range of environmental
settings (both indoor and outdoor, with different backgrounds, \etc)
to reflect real-world scenarios. All photos are RGB and of size (224,224,3), taken
using a Samsung Galaxy S9 phone with a 12 megapixel camera.

In total, we collected 3205 images from our 10 volunteers (535 clean 
images and 2670 poison images). Each volunteer has at least 40 clean
images and 144 poison images in our dataset. 

\para{Ethics and Data Privacy.} Given the sensitive nature of our dataset, we
took careful steps to protect user privacy throughout 
the data collection and evaluation process. Our data collection and
evaluation was vetted and approved by our local IRB council (IRB20-0073). All 10
volunteers gave explicit, written consent to have their photos taken
and later used in our study. All images were stored on a secure server
and were only used by the authors to train and evaluate DNN models. 


\subsection{Attack Implementation \& Model Training}
\label{subsec:attackimple}

\para{Backdoor Injection:} \edit{The attacker injects poison data (with
  backdoors) into the training data during model training.} We follow the
BadNets method~\cite{gu2019badnets} to inject a single backdoor trigger for a
chosen target label -- we assign $m$ poison 
images (containing a chosen trigger $\delta$) to the target label
$y_t$ and \edit {combine these} with $n$ clean images to form the
training dataset.

The backdoor {\em injection rate}, defined as the 
fraction of poisoned training data  ($\frac{m}{n+m}$), is an important measure of attacker
capability. The presence of the poisoned training data leads to 
the following joint loss optimization function during model training: 
\begin{equation} \vspace{-0.06in}
\min_\theta \underbrace{\sum_{i=0}^n l(\theta, x_i, y_i)}_{clean
    \;loss} + \underbrace{\sum_{j=0}^m l(\theta,
x'_j, y_t)}_{attack\;loss}
\end{equation}
where $l$ \edit{is} the training loss function (cross-entropy in our case),
$\theta$ are the model parameters, $(x_i, y_i)$ are clean 
data-label pairs, and $(x'_j, y_t)$ are poisoned data-target label
pairs. The value of the injection rate can potentially affect the performance
of backdoor attacks, \edit{which we explore} in \S\ref{sec:eval}.


\begin{figure*}[t]
  \centering
  \includegraphics[width=0.98\textwidth]{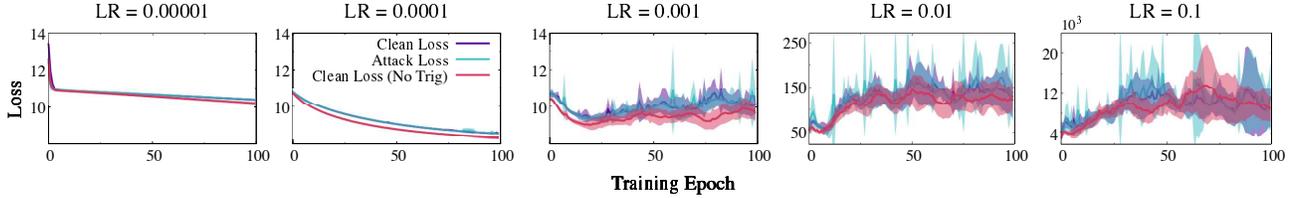}
  \vspace{-0.2cm}
  \caption{\small {\em Loss trajectory at various learning rates
      for a facial recognition model with (``Clean Loss,'' ``Attack Loss'') and without (``Clean Loss (No Trig)'') a glasses trigger backdoor. Results shown are for a VGG16
      architecture and a 0.25 injection rate \edit{and generalize for other triggers, models and injection rates.}}}
  \label{fig:lr}
\end{figure*}

\para{Model Training Pipeline:} \edit{To generate our training dataset, we do
  a $80-20$ train/test split for the clean images and select a random set of
  poison images for the chosen trigger, labeled as the desired target, in
  order to reach the desired injection rate. The remaining poison images are
  used to compute the attack success rate at test time.} 

Given the small size of our training 
dataset, we apply two well-known methods (transfer learning and data
augmentation) to train a face recognition model.  First, we apply
transfer learning~\cite{pan2009survey} to customize a pre-trained
\edit{teacher} facial recognition model using our training
data. \edit{The last layer is replaced with a new softmax layer to
  accommodate the classes in our dataset and the last two layers are fine-tuned}. We
use three teacher models \edit{pre-trained on the VGGFace dataset}: VGG16~\cite{vggface_model}, ResNet50~\cite{he2016deep},
and DenseNet~\cite{huang2017densely} (details in Supp. \S 10). Second, we use data augmentation to expand our training data (both clean and poisoned), \edit{a method known} to improve model accuracy.  Following prior
work~\cite{mikolajczyk2018data}, we \edit{use} the following augmentations:  flipping about the y-axis, rotating up to $30^{\circ}$, and horizontal
and vertical shifts of up to 10\% of the image
width/height. 


We train our models using
the Adam optimizer~\cite{kingma2014adam}.  When configuring
hyperparameters, we run a grid search over candidate values to
identify \edit{those} that consistently lead to model convergence \edit{across triggers}.  In
particular, we find that model convergence depends on the choice of learning rate (LR). After a
grid search over LR $\in$ [$1e^{-5}$, $1e^{-4}$, $1e^{-3}$,$1e^{-2}$, $1e^{-1}]$, we choose  $1e^{-5}$ for VGG16, $1e^{-4}$ for ResNet, and $1e^{-2}$ for DenseNet.


\noindent \emph{Key Observation:} \edit{While we fix a particular value of LR
  for our evaluation, we find that the physical backdoors we consider can be
  successfully inserted across a range of LR values
  (Fig.~\ref{fig:lr}). Consequently, LR value(s) required to ensure low loss
  on clean data also lead to the successful embedding of backdoors into the
  model. Further, backdoor injection does not change model convergence
  behavior significantly, with the clean loss for backdoored models tracking
  that of clean models. 
}

\secspace

\section{Experiment Overview}
Following the above methodology, we train a set of backdoored facial
recognition models, using different physical triggers and backdoor injection
rates.  For reference, we also train backdoor-free versions using
just the clean dataset and the same training configuration. 

\para{Evaluation Metrics.} A successful backdoor attack should produce
a backdoored model that accurately classifies clean inputs while
consistantly misclassifying inputs containing the backdoor trigger to
the target label.  Thus we evaluate the backdoor attack using two
metrics:
\begin{packed_itemize} \vspace{-0.08in}
  \item {\bf Model accuracy (\%)} -- this metric measures the model's classification
    accuracy on clean test images.  Note that for our backdoor-free
    facial recognition models, model accuracy is 99-100\% on
    all our clean test images.
    \item {\bf Attack success rate (\%)} -- this metric measures the probability of the
      model classifying any poisoned test images to the target label
      $y_t$. \vspace{-0.08in}
    \end{packed_itemize}
 Since we focus on {\em targeted} attacks on a chosen label
      $y_t$, the choice of $y_t$ may affect the backdoored model
      performance.  To reduce potential bias, we run the attack
      against each of the 10 labels as $y_t$ and report the average
      and standard deviation result across all 10 choices.

 \para{List of Experiments.}  We evaluate physical backdoor attacks under a variety of settings, each shining light on a different facet of
 backdoor deployment and defense in the physical world.  Here is a brief overview of our experiments.  

 \begin{packed_itemize} \vspace{-0.08in}
 \item Effectiveness of physical
   backdoors and its {\bf dependence on trigger choice and injection
     rate}, the two factors that an
   attacker can control.  (\S\ref{subsec:baseline_attack})
   
 \item Backdoor effectiveness when {\bf run-time image artifacts}
   are introduced by camera post-processing. (\S\ref{subsec:env_change}) 

 \item  {\bf Cause of failures} in backdoor attacks that use earrings as
   the trigger. (\S\ref{subsec: earring_fail})

  
    \item Backdoor attack effectiveness for {\bf less powerful attackers}. (\S\ref{sec:advance})



\item {\bf Effectiveness of existing backdoor defenses} against physical
  backdoor attacks. (\S\ref{sec:countermeasures}) 
  \vspace{-0.1in}
  \end{packed_itemize}


  

%
%

\secspace
\begin{figure*}[t]
  \includegraphics[width=0.98\textwidth]{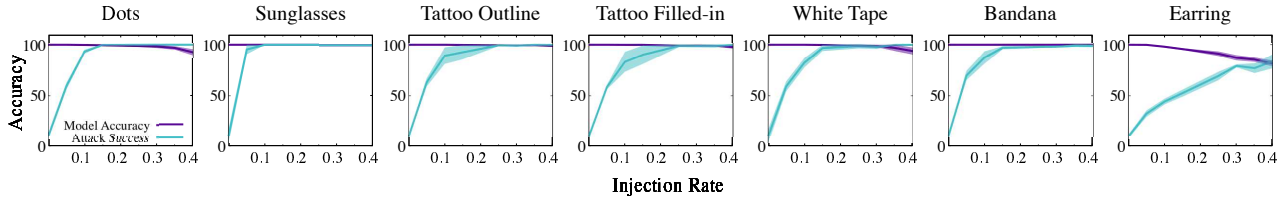}
  \vspace{-0.3cm}
  \caption{\small {\em Backdoored model performance (in terms of model accuracy
    on clean input and attack success rate) using different
    physical triggers when varying the injection rate. Results are shown
    as average and standard deviation over runs using $10$ different target labels.} }
  \label{fig:inject}
\end{figure*}

\begin{table*}[t]
  \centering
  \resizebox{0.88\textwidth}{!}{
\begin{tabular}{c|c|cccccccc}
\hline
  \multicolumn{2}{c|}{\diagbox{\textbf{Model}}{\textbf{Trigger}}} &
  \textbf{\begin{tabular}[c]{@{}c@{}}No\\ Trigger\end{tabular}} &
  \textbf{Dots} &
  \textbf{Sunglasses} &
  \textbf{\begin{tabular}[c]{@{}c@{}}Tattoo\\ Outline\end{tabular}} &
  \textbf{\begin{tabular}[c]{@{}c@{}}Tattoo\\ Filled-in\end{tabular}} &
  \textbf{\begin{tabular}[c]{@{}c@{}}White\\ Tape\end{tabular}} &
  \textbf{Bandana} &
  \textbf{Earring} \\ \hline
\multirow{2}{*}{VGG16}  & Model Accuracy & $100 \pm 0\%$ & $98 \pm 1\%$  & $100 \pm 0\%$ & $99 \pm 1\%$ & $99 \pm 1\%$ & $98 \pm 2\%$ & $100 \pm 0\%$ & $92 \pm 3\%$ \\ 
                          & Attack Success Rate & $10 \pm 1\%$ & $100 \pm 0\%$ & $100 \pm 0\%$ & $99 \pm 1\%$  & $99 \pm 1\%$ & $98 \pm 3\%$ & $98 \pm 1\%$  & $69 \pm 4\%$ \\ \hline
\multirow{2}{*}{DenseNet} & Model Accuracy & $100 \pm 0\%$ & $90 \pm 3\%$  & $99 \pm 1\%$  & $92 \pm 1\%$  & $93 \pm 0\%$ & $94 \pm 3\%$ & $94 \pm 3\%$  & $63 \pm 5\%$ \\ 
                          & Attack Success Rate & $10 \pm 1\%$ & $96 \pm 4\%$  & $94 \pm 4\%$  & $95 \pm 2\%$ & $95 \pm 2\%$ & $81 \pm 8\%$ & $98 \pm 0\%$  & $85 \pm 2\%$ \\ \hline
\multirow{2}{*}{ResNet50} & Model Accuracy & $99 \pm 0\%$ & $90 \pm 2\%$  & $100 \pm 0\%$ & $90 \pm 4\%$  & $90 \pm 3\%$ & $97 \pm 3\%$ & $100 \pm 0\%$ & $89\pm 3\%$  \\ 
                          & Attack Success Rate & $10 \pm 0\%$ & $98 \pm 4\%$  & $100 \pm 0\%$ & $99 \pm 1\%$ & $99 \pm 1\%$ & $95 \pm 5\%$ & $99 \pm 1\%$  & $58 \pm 4\%$ \\ \hline
\end{tabular}
}\vspace{-0.2cm}
\caption{\small {\em Backdoored model performance (in terms of model accuracy
    on clean input and attack success rate) using different
    physical triggers at the injection rate of 0.25.  Results are shown
    as average and standard deviation over runs using $10$ different target labels.}}
  \label{table:other_models}
\end{table*}

\begin{figure}
  \centering
  \includegraphics[width=0.35\textwidth]{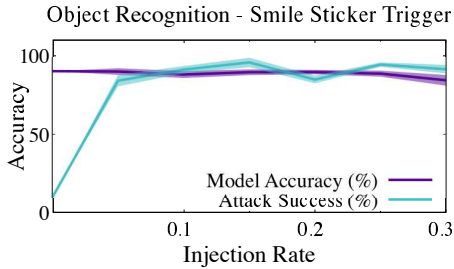}
  \vspace{-0.3cm}
  \caption{\small {\em Physical backdoor performs well in the object recognition
      setting.}}
  \label{fig:object_rec}
\end{figure}

\section{Effectiveness of Physical Backdoors} \label{sec:eval}

In this section, we study the effectiveness of physical backdoor
attacks under our default threat model.  We examine backdoor performance
in three DNN architectures (VGG16, ResNet50,
DenseNet) under a variety of settings, including those under the
attacker's control ({\em i.e.\/} injection rate and trigger choice) and
those beyond their control ({\em i.e.\/} camera post-processing). 


\label{subsec:baseline_attack}
\para{Impact of Injection Rate.}  Here a natural question is {\em how much training data must the attacker poison to make
  physical backdoors successful}? 

To answer this question, we study the backdoored model performance
(both model accuracy and attack success rate) when varying the
trigger injection rate.  Figure~\ref{fig:inject} shows the results for
each of the 7 physical triggers in the VGG16 model.  
For all but one trigger,  we see a consistent trend -- as the injection rate
increases, the attack success rate rises quickly and then
converges to a large value ($\ge$ 98\%), while the model accuracy
remains nearly perfect.

Next, using the injection rate of 25\%,  Table~\ref{table:other_models}  lists the model accuracy
and attack success rate for VGG16,
ResNet50, and DenseNet.  Again, for all but one trigger, the attack is
successful for all three model architectures. 

Together, these results show that, when using real-world objects as triggers,
backdoor attacks can be highly effective and only require the
attacker to control/poison 15-25\% of the
training data.  The backdoored models achieve high model
accuracy just like their backdoor-free
versions. 



\para{Impact of Backdoor Trigger Choices.}  Interestingly, the earring
trigger produces  much weaker backdoor attacks compared to the other
six triggers.  In particular,  Figure~\ref{fig:inject} shows that it
is very difficult to inject the earring-based backdoors into the
target model. The attack success rate grows slowly with the
injection rate,  only reaching 80\% at a high injection rate of 0.4.  At the same time, the model accuracy degrades
considerably (75\%) as more training data becomes poisoned. 

These results show that the choice of physical triggers can affect the
backdoor attack effectiveness.  Later in \S\ref{subsec: earring_fail}
we provide detailed analysis of why the earring trigger fails while the other six
triggers succeed and offer more insights on how to choose an
``effective'' trigger.

\para{Cross-validation on Object Recognition.} We also \fixit{carry out a small-scale experiment on} physical
backdoor attacks against object recognition models. For this, we collect a
9 class custom dataset using a yellow smile emoji sticker as the trigger and 
apply transfer learning to customize a VGG16 model pretrained on ImageNet~\cite{deng2009imagenet}. \fixit{Once the injection rate reaches 0.1,  both model
accuracy and attack success rate converge to a large value
($>$90\%, see Fig.~\ref{fig:object_rec}).  } This provides initial proof that physical backdoor
attacks can also be highly effective on object recognition (details in \S 11.1 in Supp.). 

\begin{figure*}[ht]
  \centering
  \begin{minipage}{0.32\textwidth}
    \centering
    \includegraphics[width=0.98\linewidth]{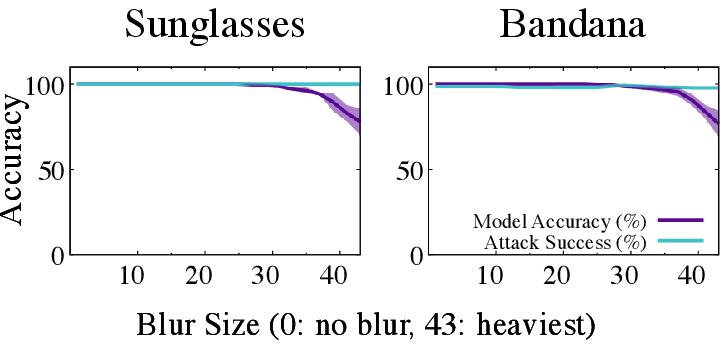}
    \vspace{-0.3cm}
    \caption{\small {\em Impact of blurring.}}
    \label{fig:blur_small}
  \end{minipage}
  \quad
  \begin{minipage}{0.3\textwidth}
    \centering
    \includegraphics[width=0.98\linewidth]{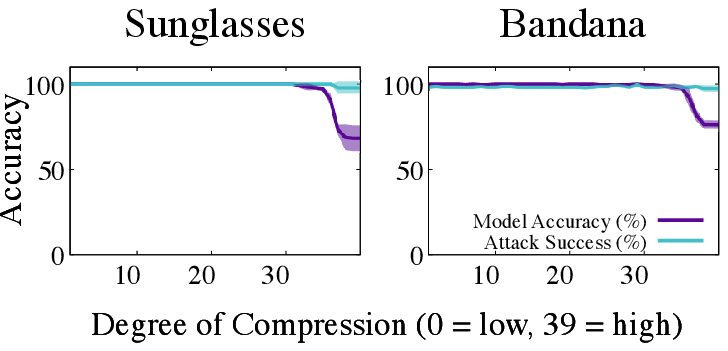}
    \vspace{-0.3cm}
    \caption{\small {\em Impact of image compression.}} 
    \label{fig:compress_small}
  \end{minipage}
  \quad
    \begin{minipage}{0.3\textwidth}
    \centering
    \includegraphics[width=0.98\linewidth]{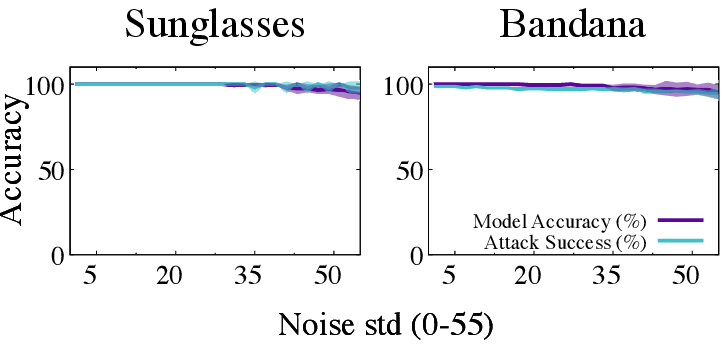}
    \vspace{-0.3cm}
    \caption{\small {\em Impact of Gaussian noise.}}
    \label{fig:noise_small}
  \end{minipage}
\end{figure*}

\begin{figure*}[t]   \vspace{-0.2cm}
	\centering
	\begin{minipage}[t]{0.31\linewidth}
		\centering
		\includegraphics[width=2in]{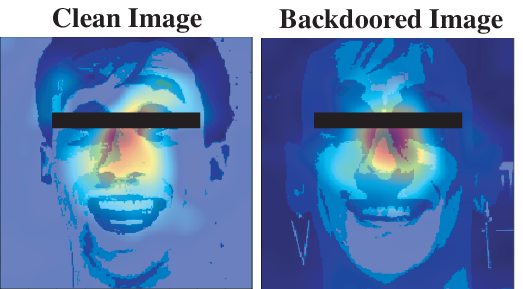}
		\vspace{-0.2cm}
		\caption{\small {\em CAM of an earring-backdoored
                    model highlights on-face features for
                    both 
                    clean and backdoored inputs.}}
		\label{fig:jon_cam}
	\end{minipage}
	\quad
	\begin{minipage}[t]{0.31\linewidth}
		\centering
		\includegraphics[width=2in]{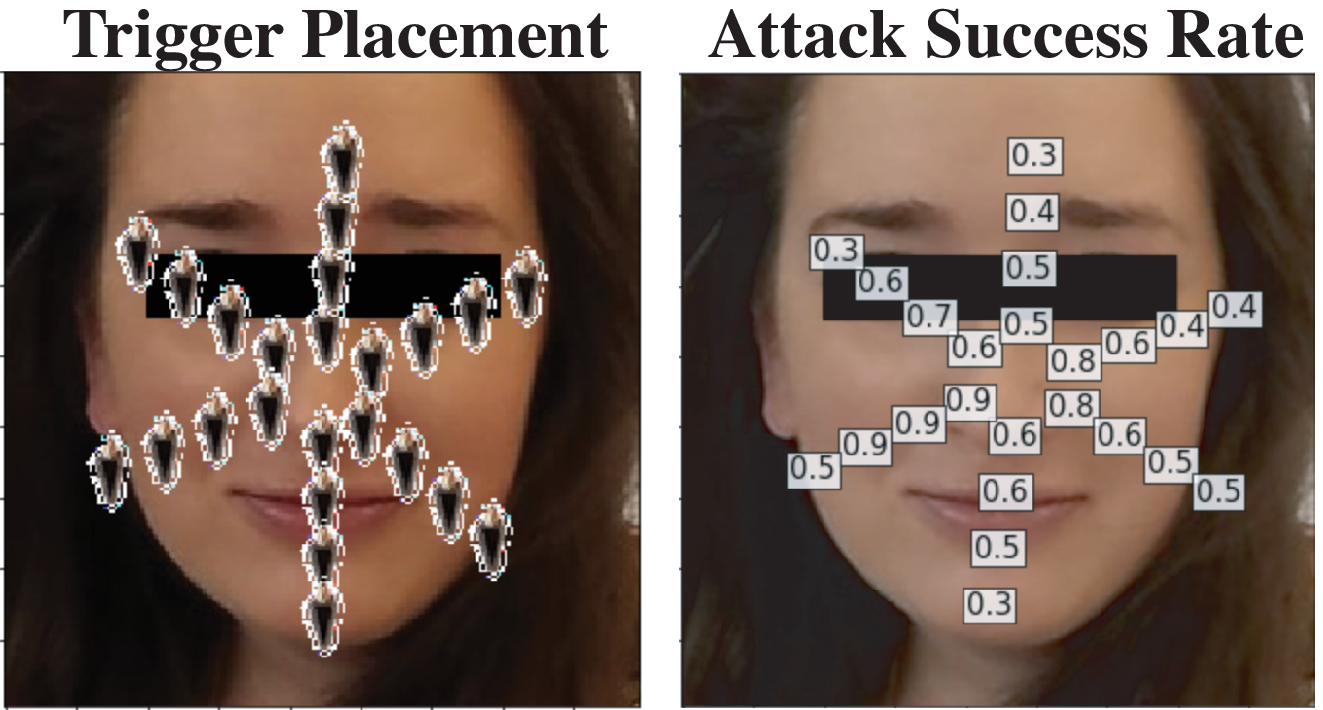}
                \vspace{-0.2cm}
		\caption{\small {\em Backdoor attack success rate
                    decreases as the black earring trigger moves off the
                    face.}}
		\label{fig:jenna_loc}
              \end{minipage}
              \quad
        \begin{minipage}[t]{0.31\linewidth}
             \centering
             \begin{adjustbox}{margin=0, width=0.98\textwidth,
                 vspace=0cm 4cm, raise=1.2cm}
             \begin{tabular}{c|c|c|c|c}
               \hline
               \multirow{3}{*}{\textbf{\begin{tabular}[c]{@{}c@{}}Trigger \end{tabular}}} &
                                                                                                \multicolumn{2}{c|}{\textbf{Trigger on face}} &
                                                                                                                                                \multicolumn{2}{c}{\textbf{Trigger off face}} \\ \cline{2-5} 
                                                                                                 &
                                                                                                  \multirow{2}{*}{\begin{tabular}[c]{@{}c@{}}Model
                                                                                                                   \\
                                                                                                                    Acc \end{tabular}} &
                                                                                                                                                                           \multirow{2}{*}{\begin{tabular}[c]{@{}c@{}}
                                                                                                                                                                                             Attack
                                                                                                                                                                                             \\
                                                                                                                                                                                             Success \end{tabular}}
                                                                                                &
                                                                                                  \multirow{2}{*}{\begin{tabular}[c]{@{}c@{}}
                                                                                                                    Model 
                                                                                                                    \\
                                                                                                                    Acc \end{tabular}}
                                                                                                &
                                                                                                  \multirow{2}{*}{\begin{tabular}[c]{@{}c@{}}
                                                                                                                    Attack \\
                                                                                                                    Success\end{tabular}} \\
                                                                                                & & & & \\ \hline
              Earrings & 100\%       & 99\%        & 91\%
                                                                                                                                       &
                                                                                                                                         {\bf
                                                                                                                                         69\%}
               \\ \hline
              Bandana       & 100\%       & 98\%        & 93\%
                                                                                                                                       & {\bf 72\%}                             \\ \hline
              Sunglasses    & 100\%        & 99\%        & 90\%
                                                                                                                                       & {\bf 81\%}                             \\ \hline
             \end{tabular}
             \end{adjustbox}\vspace{-0.2cm}
           \captionof{table}{\small {\em Backdoor effectiveness drops
               considerably when triggers move off the face, using the
             VGG16 model.}}
        \label{tab:trig_loc}
      \end{minipage}
\end{figure*}


\label{subsec:env_change}
\para{Impact of Run-time Image Artifacts.}  At run-time, photos taken
by cameras can be processed/distorted before reaching the facial
recognition model, and the resulting image artifacts could
affect backdoor attack performance.  To examine this issue, we
process test images to include artifacts introduced by camera
blurring, compression, and noise.  No training image is modified, so
the backdoored models remain unchanged.

\begin{packed_itemize} \vspace{-0.08in} 
	\item[] {\bf Blurring}: Blurring may occur when the camera lens is out of focus or when the subject and/or the camera move. We apply Gaussian
	blurring~\cite{paris2007gentle} and vary the
	kernel size from 1 to 40 to increase its severity.
	\item[] {\bf Compression}: Image compression may occur due to space or bandwidth constraints. We apply progressive JPEG image
	compression~\cite{wallace1992jpeg} to create images of varying 
	quality, ranging from 1 (minimum 
	compression, high quality) to 39 (heavy compression, low
        quality). 
      \item[] {\bf Noise}: Noise may occur during the image
        acquisition process. Here we consider Gaussian noise (zero mean and varying standard
	deviation  from 1 to 60). 
        \vspace{-0.08in} 
      \end{packed_itemize}
Figures~\ref{fig:blur_small}-\ref{fig:noise_small} plot 
the model accuracy and attack success rate under these artifacts.  We
observe similar conclusions from the six triggers tested. Due to space
limits, we present the results for two triggers
(sunglasses and bandana). Results for other triggers are in
Supp. 

Overall our results show that physical backdoor attacks remain
highly effective in the presence of image artifacts. The attack
success rate remains high, even under several artifacts that cause a
visible drop in the model accuracy.  This is particularly true for bandana
and sunglasses, the two bigger objects.   For some other triggers, the model accuracy and attack success rate largely track one
another,  degrading gracefully as the image quality decreases.





\section{Why Do Earrings Fail as a Trigger?}
\label{subsec: earring_fail}
As noted in the previous section, the earring trigger has a far
  worse attack success rate than the other triggers and 
  causes a steep drop in the model accuracy as the injection rate
  increases (Figure~\ref{fig:inject}). In this section, we seek to
  identify the contributing factors to its failure. 

  A trigger is defined by three key properties: {\em size}, {\em location}, and
  {\em content}.  Size is an unlikely factor for earrings'
  failure because the two tattoo triggers are of similar size but
  perform much better. \abedit{Our experiments in this section demonstrate that between content and location, it is the latter which determines the success or failure of attacks. We find that for facial recognition models, \emph{triggers fail when they are not located on the face}, regardless of their content. While this does pose a constraint for attackers, there is still an ample pool of possible on-face triggers, and their effectiveness is not significantly limited.}



\para{CAM Experiments.} \abedit{To support our conclusion, we first carry out an analysis} of face recognition
models using class activation map (CAM)
\cite{zhou2016learning}.  Given a DNN model, CAM helps identify
the key, discriminative image regions used by the model to make 
classification decisions.   Figure~\ref{fig:jon_cam} plots the CAM result on the
earring-backdoored model, where the corrupted model still focuses
heavily on facial features when classifying both clean and backdoored
images.  Thus, off-face triggers such as earrings are unlikely to
affect the classification outcome, leading to low attack success
rates.  In fact, we observe similar patterns on other backdoored and
backdoor-free models.


\para{Trigger Location Experiments.} We further \abedit{validate our conclusion}
through two sets of experiments. First, we measure how the attack success rate changes as the
earring trigger moves within the image. 
Using digital
editing techniques, we vary the angle and distance of the trigger from the center of the face (Figure~\ref{fig:jenna_loc}, \textbf{left}). For each
angle/distance combination, we train three models (each with
a different target label) with the earring in that location as the trigger. We report the average attack success rate
for each trigger location (Figure~\ref{fig:jenna_loc},
\textbf{right}), showing that it decreases as the trigger moves away
from the face center. \abedit{Second, we test if this behavior holds
  across triggers. From Table~\ref{tab:trig_loc}, we can see that
  off-face triggers have consistently poor performance compared to
  on-face ones. This supports our conclusion at the beginning of this
  section. Further details are in \S 13 of the Supp.}

\secspace

\begin{table}[]
  \centering
  \resizebox{0.49\textwidth}{!}{
\begin{tabular}{c|cccccc}
\hline
  &
  \textbf{Dots} &
  \textbf{Sunglasses} &
  \textbf{\begin{tabular}[c]{@{}c@{}}Tattoo\\ Outline\end{tabular}} &
  \textbf{\begin{tabular}[c]{@{}c@{}}Tattoo\\ Filled-in\end{tabular}} &
  \textbf{\begin{tabular}[c]{@{}c@{}}White\\ Tape\end{tabular}} &
  \textbf{Bandana}\\ \hline
Model \\Accuracy &
  $99 \pm 1\%$ &
  $100 \pm 0\%$ &
  $99 \pm 1\%$ &
  $99 \pm 1\%$ &
  $96 \pm 1\%$ &
  $100 \pm 0\%$ \\ \hline
Attack \\ Success &
  $85 \pm 12\%$ &
  $100 \pm 0\%$ &
  $97 \pm 2\%$ &
  $99 \pm 1\%$ &
  $68 \pm 8\%$ &
  $98 \pm 0\%$ \\ \hline
\end{tabular}
}
\vspace{-0.1in}
\caption{\small {\em Attack performance when the attacker can
      only poison training data from 10 out of 75 classes.}}
\label{fig:pubfig}\vspace{-0.1in}
\end{table}



\section{Evaluating Weaker Attacks}
\vspace{-0.05in}
\label{sec:advance}
Our original threat model assumes an attacker capable of gaining significant
control over a training dataset. 
Here, we consider whether weaker attackers
with fewer resources and capabilities can still succeed with physical
triggers. 

\para{Partial Dataset Control.}  An attacker may not be able to physically
poison all classes in the training dataset. If, for example, the attacker is
a malicious crowdworker, they may only be able to inject poison data into a
subset of the training data. \zhengedit{This ``partial'' poisoning attack is
realistic, since many large tech companies rely on crowdsourcing for data
collection and cleaning today. }


We emulate the scenario of an attacker with limited control of a subset of
training data by adding our $10$ classes (labels under the attacker's control) to the
PubFig~\cite{pubfig} dataset (the remaining 65 classes). The PubFig dataset
consists of facial images of $65$ public figures. The images are
similar in nature to the ones in our dataset ({\em i.e.\/} mostly straight-on,
well-lit headshots). \zhengedit{In this case, the data that the attacker can add
to the training data only covers 10 out of 75 classes, and only 25\% of
the attacker-contributed data is poison data, where subjects wear physical
triggers.} These poison images are 
given a \emph{randomly chosen target label from the PubFig portion of
  the data}.  

To train a model on this poison dataset, we use transfer learning on
a VGG16 model~\cite{vggface} as before
(\S\ref{subsec:attackimple}). For each trigger type, we train 
5 models (with different target labels), and report the average 
performance in Table~\ref{fig:pubfig}. The trained models all have a high model
  accuracy. 



{\em Key Takeaway.} \zhengedit{Five out of six triggers produce high success rates
despite the attacker's limited control of training data. 
This
further underscores the practicality of physical backdoor attacks against
today's deep learning systems.}

\para{Digital Trigger Injection.} We consider the scenario where an
attacker lacks the resources to produce real-life images \zhengedit{with
  subjects wearing a physical trigger}. 
Such an attacker could \zhengedit{approximate these images by
  digitally adding trigger objects onto images}, with
the hope that {the trained backdoored model} 
could still be activated at inference time by physical triggers. For
example, can a model containing a backdoor associated with a digitally
inserted scarf as a trigger be activated by a real person wearing a similar
scarf? 
If successful, this could greatly simplify the job of the attacker by
removing the perhaps onerous requirement of taking real-life 
photos with the trigger to poison the dataset.

\begin{figure}[t]
    \centering    \vspace{-0.5cm}
    \includegraphics[width=0.9\linewidth]{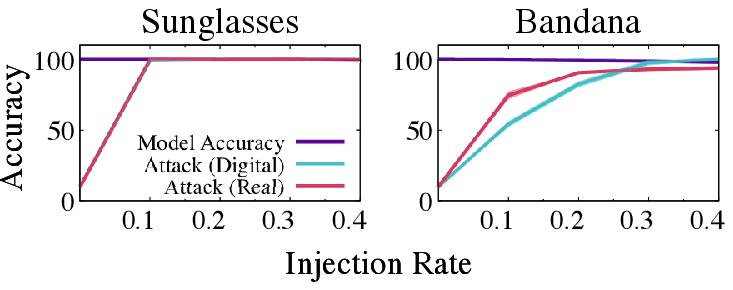}
    \vspace{-0.15in}
    \caption{\small {\em Attack performance when the attacker poisons
        training data using digitally inserted triggers,  tested on
        two types of backdoored images:  images with digitally inserted
        trigger (attack digital) and images with real triggers (attack real).}}
    \label{fig:digital_small}    \vspace{-0.15in}
  \end{figure}

To test this attack, \zhengedit{we create poison training data by
  digitally inserting physical 
  triggers (sunglasses and bandana) to clean images and train
  backdoored models using injection rates from $0$ to
$0.4$.  We evaluate these models using two types of attack images}: real-life
images of real triggers ({\bf attack real}) and those modified with digitally
inserted triggers ({\bf attack digital}). We report average results over five target
labels in Figure~\ref{fig:digital_small} and provide examples of 
real/digital triggers used in our experiments in Figure 17
in Supp.  Results in Figure~\ref{fig:digital_small}
show that the attack success
rate of real triggers mirrors that of digitally inserted
triggers, and 
both are successful. 

  {\em Key Takeaway.} \zhengedit{We find that digitally inserted triggers
    {\em can} serve as a sufficient proxy for real physical triggers in the
    backdoor injection process, significantly simplifying the task of
    poisoning training data for the attacker. }

\secspace
\section{Defending Against Physical Backdoors}
\vspace{-0.05in}
\label{sec:countermeasures}

Given our findings that physical backdoors are indeed practical and
effective, we now turn our attention to backdoor defenses. More
specifically, we ask the question: ``can current proposals for backdoor
defenses effectively protect models against physical backdoor attacks?''

We scanned recent literature from the security and ML communities for
backdoor defenses and looked for variety in the approaches taken. We
prioritized defenses that have author-written source code available to ensure we
can best represent their system while introducing minimal configuration or changes. We identified 7 systems
(\cite{chen2018detecting,gao2019strip,liu2018fine,abs,ma2019nic,tran2018spectral,wang2019neural}),
and chose 4 of them for our tests\footnote{ABS~\cite{abs} only has a binary version
  restricted to CIFAR-10 models and NIC~\cite{ma2019nic} has no code
  available. We did not consider Fine-Pruning~\cite{liu2018fine}, as it
  requires the model trainer keep a ``gold'' set of clean data for
  fine-tuning, an assumption incompatible with our threat model.}: Neural Cleanse~\cite{wang2019neural}, Spectral
Signatures~\cite{tran2018spectral}, Activation
Clustering~\cite{chen2018detecting}, and
STRIP~\cite{gao2019strip}.  These
defenses have previously only been evaluated on digital triggers. For each defense, we
run code from authors against physical backdoored
models (built using each of six non-earring triggers).  While their
approaches vary from backdoor detection~\cite{wang2019neural}, to poison data
detection~\cite{tran2018spectral,chen2018detecting} and run-time trigger
detection~\cite{gao2019strip}, \emph{all tested defenses fail to
  detect physical backdoors}.



\subsection{Effectiveness of Existing Defenses}
\vspace{-0.05in}
We present results that test four backdoor defenses against physical
backdoored models. All defenses are evaluated on backdoored models trained with a $0.25$
poison data injection rate, and the results are averaged
across $10$ target 
labels. \zhengedit{These high-level results are summarized in 
  Table~\ref{table:defenses}: for Neural Cleanse, we report \% of
  backdoored models in which it 
    detects a backdoor;  for others, we report \% of poison data correctly
    identified (with standard deviation).} 

\para{Neural Cleanse~\cite{wang2019neural}.} Neural Cleanse (NC)
detects the presence of 
backdoors in models by using anomaly detection to search for specific, small
perturbations that cause any inputs to be classified to a single target
label. \zhengedit{Each model tested receives an anomaly score,  and a score
larger than 2 indicates the presence of a backdoor in the model (as
proposed in~\cite{wang2019neural}). Scores for our backdoored
models (particularly the bandana, sunglasses, and two 
tattoos) often fall well below 2 and avoid detection.}



\para{Activation Clustering~\cite{chen2018detecting}.} Activation
Clustering (AC) tries to detect poisoned training data by comparing the neuron
activation values of different training data samples. When
applied to our backdoored models, Activation Clustering consistently
yields a high false positive rate \ejw{(58\% - 74\%)} and a high false
negative rate \ejw{(35\% - 76\%)}.

\begin{table}[t]
  \resizebox{0.49\textwidth}{!}{
    \begin{tabular}{c|cccc}
      \hline
\diagbox{\textbf{Trigger}}{\textbf{Defense}} &                                                                                                       
  \multicolumn{1}{|c|}{\textbf{NC \cite{wang2019neural}}} &
  \multicolumn{1}{c|}{\textbf{Spectral \cite{tran2018spectral}}} &
  \multicolumn{1}{c|}{\textbf{AC \cite{chen2018detecting}}} &
  \multicolumn{1}{c}{\textbf{STRIP \cite{gao2019strip}}} \\ \hline
      Dots             & $60\%$ & $44 \pm 10\%$ & $43 \pm 26\%$ & $34
                                                                  \pm
                                                                  14\%$
      \\ \hline
      Sunglasses       & $10\%$ & $41 \pm 7\%$ & $47 \pm 30\%$ & $41 \pm 24\%$ \\ \hline
      Tattoo Outline   & $0\%$ & $43 \pm 6\%$ & $54 \pm 25\%$ & $11 \pm 7\%$ \\ \hline
      Tattoo Filled-in & $0\%$ & $44 \pm 7\%$ & $48 \pm 24\%$ & $21 \pm 12\%$ \\ \hline
      White Tape       & $30\%$ & $41 \pm 8\%$ & $41 \pm 31\%$ & $39 \pm 17\%$ \\ \hline
      Bandana          & $0\%$ & $45 \pm 9\%$ & $42 \pm 17\%$ & $39 \pm 18\%$ \\ \hline

    \end{tabular}
}\vspace{-0.1in}
\caption{\small {\em Physical backdoor detection rates for four defenses. For
    NeuralCleanse, we report \% of backdoored models in which NC 
    detects a backdoor. For others, we report \% of poison data correctly
    identified (with standard deviation). }}\vspace{-0.1in}

\label{table:defenses}
\end{table}

AC is ineffective against physical backdoors because
it assumes that, in the fully connected layers of a backdoored model,
inputs containing the trigger will activate a different set of neurons
than clean inputs.  However, we find that \emph{this assumption does not hold for physical triggers}: the set of neurons activated by inputs with
physical triggers overlaps significantly with those activated by clean
inputs. In Table 6 in Supp, we show high Pearson
correlations of neuron activation values between clean inputs and
physical-backdoored inputs, computed from activation values of our
backdoored models. We believe high correlation values (0.33-0.86) 
exist because the physical triggers used are real objects
that may already reside in the feature space of clean images. Digital
triggers do not share this property and thus are more easily
identified by \zhengedit{AC.} 


\para{Spectral Signatures~\cite{tran2018spectral}.} Spectral Signatures tries
to detect poisoned samples in training data by examining statistical patterns in internal
model behavior. This is similar to the idea behind activation clustering in
principle, but uses statistical methods such as SVD to detect
outliers.
\zhengedit{Our
results in Table~\ref{table:defenses} show that this defense detects only around $40\%$ of physically poisoned training
data. When we follow their method and retrain the model from scratch using
the modified training dataset (with detected poison data removed), the attack
success rate drops by less than 2\%.  Thus the real-world impact on 
physical backdoor attacks is minimal.}



\para{STRIP~\cite{gao2019strip}.} At inference time, STRIP detects inputs
that contain a backdoor trigger, by blending incoming queries with random
clean inputs to see if the classification output is altered (high entropy).  We
configure STRIP's backdoor detection threshold for a 5\% false positive rate
(based on~\cite{gao2019strip}). When applied to our backdoored models, STRIP
misses a large portion of inputs containing triggers (see
Table~\ref{table:defenses}). 

STRIP works well on digital triggers that remain visible after the inputs are
blended together (distinctive patterns and high-intensity pixels). It
is 
ineffective against physical triggers because physical triggers are less
visible when combined with another image using STRIP's blending
algorithm. Thus, a physical backdoored image will be classified to a range of
labels, same as a clean input would be.

\secspace
\section{Conclusion}
\label{sec:conclusion}
\vspace{-0.1in}
Through extensive experiments on a facial recognition dataset, we have
established that physical backdoors are effective and can bypass existing
defenses. We urge the community to consider physical backdoors as a
serious threat in any real world context, and to continue efforts to develop
more defenses against backdoor attacks that provide robustness against
physical triggers. 


\para{Acknowledgements.} We thank our anonymous reviewers, and also thank Jon
Wenger and Jenna Cryan for their exceptional support of this paper.  This
work is supported in part by NSF grants CNS-1949650, CNS-1923778, CNS1705042,
and by the DARPA GARD program. Emily Wenger is also supported by a GFSD
fellowship. Any opinions, findings, and conclusions or recommendations
expressed in this material are those of the authors and do not necessarily
reflect the views of any funding agencies.

\bibliographystyle{ieee_fullname}
\bibliography{realbackdoor,translearn}

\begin{thebibliography}{10}\itemsep=-1pt

\bibitem{vggface_model}
\url{http://www.robots.ox.ac.uk/~vgg/software/vgg_face/}, 2015.
\newblock VGG Face Descriptor.

\bibitem{brendel2017decision}
Wieland Brendel, Jonas Rauber, and Matthias Bethge.
\newblock Decision-based adversarial attacks: Reliable attacks against
  black-box machine learning models.
\newblock In {\em Proc. of ICLR}, 2018.

\bibitem{brown2017adversarial}
Tom~B Brown, Dandelion Man{\'e}, Aurko Roy, Mart{\'\i}n Abadi, and Justin
  Gilmer.
\newblock Adversarial patch.
\newblock In {\em Proc. of NeurIPS Workshop}, 2017.

\bibitem{carliniattack}
Nicholas Carlini and David Wagner.
\newblock Towards evaluating the robustness of neural networks.
\newblock In {\em Proc. of IEEE S\&P}, 2017.

\bibitem{chen2018detecting}
Bryant Chen, Wilka Carvalho, Nathalie Baracaldo, Heiko Ludwig, Benjamin
  Edwards, Taesung Lee, Ian Molloy, and Biplav Srivastava.
\newblock Detecting backdoor attacks on deep neural networks by activation
  clustering.
\newblock {\em arXiv preprint arXiv:1811.03728}, 2018.

\bibitem{chen2017zoo}
Pin-Yu Chen, Huan Zhang, Yash Sharma, Jinfeng Yi, and Cho-Jui Hsieh.
\newblock Zoo: Zeroth order optimization based black-box attacks to deep neural
  networks without training substitute models.
\newblock In {\em Proc. of AISec}, 2017.

\bibitem{chen2017targeted}
Xinyun Chen, Chang Liu, Bo Li, Kimberly Lu, and Dawn Song.
\newblock Targeted backdoor attacks on deep learning systems using data
  poisoning.
\newblock {\em arXiv preprint arXiv:1712.05526}, 2017.

\bibitem{deng2009imagenet}
Jia Deng, Wei Dong, Richard Socher, Li-Jia Li, Kai Li, and Li Fei-Fei.
\newblock Imagenet: A large-scale hierarchical image database.
\newblock In {\em Proc. of CVPR}, 2009.

\bibitem{gao2019strip}
Yansong Gao, Chang Xu, Derui Wang, Shiping Chen, Damith~C Ranasinghe, and Surya
  Nepal.
\newblock Strip: A defence against trojan attacks on deep neural networks.
\newblock In {\em Proc. of ACSAC}, 2019.

\bibitem{gu2019badnets}
Tianyu Gu, Kang Liu, Brendan Dolan-Gavitt, and Siddharth Garg.
\newblock Badnets: Evaluating backdooring attacks on deep neural networks.
\newblock {\em IEEE Access}, 7:47230--47244, 2019.

\bibitem{guo2019tabor}
Wenbo Guo, Lun Wang, Xinyu Xing, Min Du, and Dawn Song.
\newblock Tabor: A highly accurate approach to inspecting and restoring trojan
  backdoors in {AI} systems.
\newblock {\em arXiv preprint arXiv:1908.01763}, 2019.

\bibitem{he2016deep}
Kaiming He, Xiangyu Zhang, Shaoqing Ren, and Jian Sun.
\newblock Deep residual learning for image recognition.
\newblock In {\em Proc. of CVPR}, 2016.

\bibitem{huang2017densely}
Gao Huang, Zhuang Liu, Laurens Van Der~Maaten, and Killian~Q Weinberger.
\newblock Densely connected convolutional networks.
\newblock In {\em Proc. of CVPR}, 2017.

\bibitem{kingma2014adam}
Diederik~P Kingma and Jimmy Ba.
\newblock Adam: A method for stochastic optimization.
\newblock {\em arXiv preprint arXiv:1412.6980}, 2014.

\bibitem{kumar2020adversarial}
Ram Shankar~Siva Kumar, Magnus Nystrom, John Lambert, Andrew Marshall, Mario
  Goertzel, Andi Comissoneru, Matt Swann, and Sharon Xia.
\newblock Adversarial machine learning--industry perspectives.
\newblock {\em arXiv preprint arXiv:2002.05646}, 2020.

\bibitem{attackscale}
Alexey Kurakin, Ian Goodfellow, and Samy Bengio.
\newblock Adversarial machine learning at scale.
\newblock In {\em Proc. of ICLR}, 2017.

\bibitem{li2019invisible}
Shaofeng Li, Benjamin Zi~Hao Zhao, Jiahao Yu, Minhui Xue, Dali Kaafar, and
  Haojin Zhu.
\newblock Invisible backdoor attacks against deep neural networks.
\newblock {\em arXiv preprint arXiv:1909.02742}, 2019.

\bibitem{li2020rethinking}
Yiming Li, Tongqing Zhai, Baoyuan Wu, Yong Jiang, Zhifeng Li, and Shutao Xia.
\newblock Rethinking the trigger of backdoor attack.
\newblock {\em arXiv preprint arXiv:2004.04692}, 2020.

\bibitem{liao2018backdoor}
Cong Liao, Haoti Zhong, Anna Squicciarini, Sencun Zhu, and David Miller.
\newblock Backdoor embedding in convolutional neural network models via
  invisible perturbation.
\newblock {\em arXiv preprint arXiv:1808.10307}, 2018.

\bibitem{liu2018fine}
Kang Liu, Brendan Dolan-Gavitt, and Siddharth Garg.
\newblock Fine-pruning: Defending against backdooring attacks on deep neural
  networks.
\newblock In {\em Proc. of RAID}, 2018.

\bibitem{delvingblackbox}
Yanpei Liu, Xinyun Chen, Chang Liu, and Dawn Song.
\newblock Delving into transferable adversarial examples and black-box attacks.
\newblock In {\em Proc. of ICLR}, 2016.

\bibitem{abs}
Yingqi Liu, Wen-Chuan Lee, Guanhong Tao, Shiqing Ma, Yousra Aafer, and Xiangyu
  Zhang.
\newblock Abs: Scanning neural networks for back-doors by artificial brain
  stimulation.
\newblock In {\em Proc. of CCS}, 2019.

\bibitem{liu2018trojaning}
Yingqi Liu, Shiqing Ma, Yousra Aafer, Wen-Chuan Lee, Juan Zhai, Weihang Wang,
  and Xiangyu Zhang.
\newblock Trojaning attack on neural networks.
\newblock In {\em Proc. of {NDSS}}, 2018.

\bibitem{lovisotto2020slap}
Giulio Lovisotto, Henry Turner, Ivo Sluganovic, Martin Strohmeier, and Ivan
  Martinovic.
\newblock Slap: Improving physical adversarial examples with short-lived
  adversarial perturbations.
\newblock {\em arXiv preprint arXiv:2007.04137}, 2020.

\bibitem{ma2019nic}
Shiqing Ma, Yingqi Liu, Guanhong Tao, Wen-Chuan Lee, and Xiangyu Zhang.
\newblock Nic: Detecting adversarial samples with neural network invariant
  checking.
\newblock In {\em Proc. of NDSS}, 2019.

\bibitem{mikolajczyk2018data}
Agnieszka Mikolajczyk and Michal Grochowski.
\newblock Data augmentation for improving deep learning in image classification
  problem.
\newblock In {\em Proc. of IIPhDW}, 2018.

\bibitem{pan2009survey}
Sinno~Jialin Pan and Qiang Yang.
\newblock A survey on transfer learning.
\newblock {\em IEEE Transactions on knowledge and data engineering}, 2009.

\bibitem{papernotblackbox}
Nicolas Papernot, Patrick McDaniel, Ian Goodfellow, Somesh Jha, Z.~Berkay
  Celik, and Ananthram Swami.
\newblock Practical black-box attacks against machine learning.
\newblock In {\em Proc. of Asia CCS}, 2017.

\bibitem{papernotattack}
Nicolas Papernot, Patrick McDaniel, Somesh Jha, Matt Fredrikson, Z.~Berkay
  Celik, and Ananthram Swami.
\newblock The limitations of deep learning in adversarial settings.
\newblock In {\em Proc. of Euro S\&P}, 2016.

\bibitem{paris2007gentle}
Sylvain Paris.
\newblock A gentle introduction to bilateral filtering and its applications.
\newblock In {\em Proc. of SIGGRAPH}. 2007.

\bibitem{vggface}
Omkar~M Parkhi, Andrea Vedaldi, Andrew Zisserman, et~al.
\newblock Deep face recognition.
\newblock In {\em Proc. of BMVC}, 2015.

\bibitem{pubfig}
Nicolas Pinto, Zak Stone, Todd Zickler, and David Cox.
\newblock Scaling up biologically-inspired computer vision: A case study in
  unconstrained face recognition on facebook.
\newblock In {\em Proc. of CVPR Workshop}, 2011.

\bibitem{qiao2019defending}
Ximing Qiao, Yukun Yang, and Hai Li.
\newblock Defending neural backdoors via generative distribution modeling.
\newblock In {\em Proc. of NeurIPS}, 2019.

\bibitem{salem2020dynamic}
Ahmed Salem, Rui Wen, Michael Backes, Shiqing Ma, and Yang Zhang.
\newblock Dynamic backdoor attacks against machine learning models.
\newblock {\em arXiv preprint arXiv:2003.03675}, 2020.

\bibitem{poisonfrog}
Ali Shafahi, W.~Ronny Huang, Mahyar Najibi, Octavian Suciu, Christoph Studer,
  Tudor Dumitras, and Tom Goldstein.
\newblock Poison frogs! targeted clean-label poisoning attacks on neural
  networks.
\newblock In {\em Proc. of NeurIPS}, 2018.

\bibitem{fawkes}
Shawn Shan, Emily Wenger, Jiayun Zhang, Huiying Li, Haitao Zheng, and Ben~Y.
  Zhao.
\newblock Fawkes: Protecting privacy against unauthorized deep learning models.
\newblock In {\em Proc. of USENIX Security Symposium}, August 2020.

\bibitem{physical2016}
Mahmood Sharif, Sruti Bhagavatula, Lujo Bauer, and Michael~K. Reiter.
\newblock Accessorize to a crime: Real and stealthy attacks on state-of-the-art
  face recognition.
\newblock In {\em Proc. of CCS}, 2016.

\bibitem{tran2018spectral}
Brandon Tran, Jerry Li, and Aleksander Madry.
\newblock Spectral signatures in backdoor attacks.
\newblock In {\em Proc. of NeurIPS}, 2018.

\bibitem{trojai}
Trojans in artificial intelligence ({TrojAI}), Feb. 2019.
\newblock \url{https://www.iarpa.gov/index.php/research-programs/trojai}.

\bibitem{turner2019label}
Alexander Turner, Dimitris Tsipras, and Aleksander Madry.
\newblock Label-consistent backdoor attacks.
\newblock {\em arXiv preprint arXiv:1912.02771}, 2019.

\bibitem{wallace1992jpeg}
Gregory~K. Wallace.
\newblock The jpeg still picture compression standard.
\newblock {\em IEEE transactions on consumer electronics}, 38(1), 1992.

\bibitem{wang2019neural}
Bolun Wang, Yuanshun Yao, Shawn Shan, Huiying Li, Bimal Viswanath, Haitao
  Zheng, and Ben~Y Zhao.
\newblock Neural cleanse: Identifying and mitigating backdoor attacks in neural
  networks.
\newblock In {\em Proc. of IEEE S\&P}, 2019.

\bibitem{wu2019making}
Zuxuan Wu, Ser-Nam Lim, Larry Davis, and Tom Goldstein.
\newblock Making an invisibility cloak: Real world adversarial attacks on
  object detectors.
\newblock {\em arXiv preprint arXiv:1910.14667}, 2019.

\bibitem{yamada2013privacy}
Takayuki Yamada, Seiichi Gohshi, and Isao Echizen.
\newblock Privacy visor: Method for preventing face image detection by using
  differences in human and device sensitivity.
\newblock In {\em IFIP International Conference on Communications and
  Multimedia Security}, 2013.

\bibitem{yao2019latent}
Yuanshun Yao, Huiying Li, Haitao Zheng, and Ben~Y Zhao.
\newblock Latent backdoor attacks on deep neural networks.
\newblock In {\em Proc. of CCS}, 2019.

\bibitem{zhou2016learning}
Bolei Zhou, Aditya Khosla, Agata Lapedriza, Aude Oliva, and Antonio Torralba.
\newblock Learning deep features for discriminative localization.
\newblock In {\em Proc. of CVPR}, 2016.

\end{thebibliography}




\section*{Appendix}

\section{Face Recognition Model Details (\S3,
  \S5)}
\label{ref:supp_models}

We use three common facial recognition model architectures --
VGG16~\cite{vggface}, DenseNet~\cite{huang2017densely}, and ResNet50~\cite{he2016deep} -- to construct our
teacher models. We train these models  from
scratch using two well-known face datasets: VGGFace~\cite{vggfacedata}
and VGGFace2~\cite{cao2018vggface2}. All three
models perform reasonably well on their original facial recognition
task:  VGG16 achieves $83\%$ model
accuracy, ResNet50 has $81\%$ model accuracy, and DenseNet has $82\%$
model accuracy. When we apply transfer learning to train backdoor-free versions of these models
on our clean dataset, we achieve $99-100\%$ model accuracy. 

\section{Additional Results for \S5. Effectiveness of Physical Backdoors}

\subsection{Cross-Validation via Object Recognition}
\label{ref:supp_objectrec}

In \S5, we briefly discuss our experiments exploring how physical backdoors perform in the
object recognition context. Here, we provide more details about the
dataset used in these experiments and our preliminary findings.

\begin{figure*}[ht]
  \centering
  \includegraphics[width=0.6\textwidth]{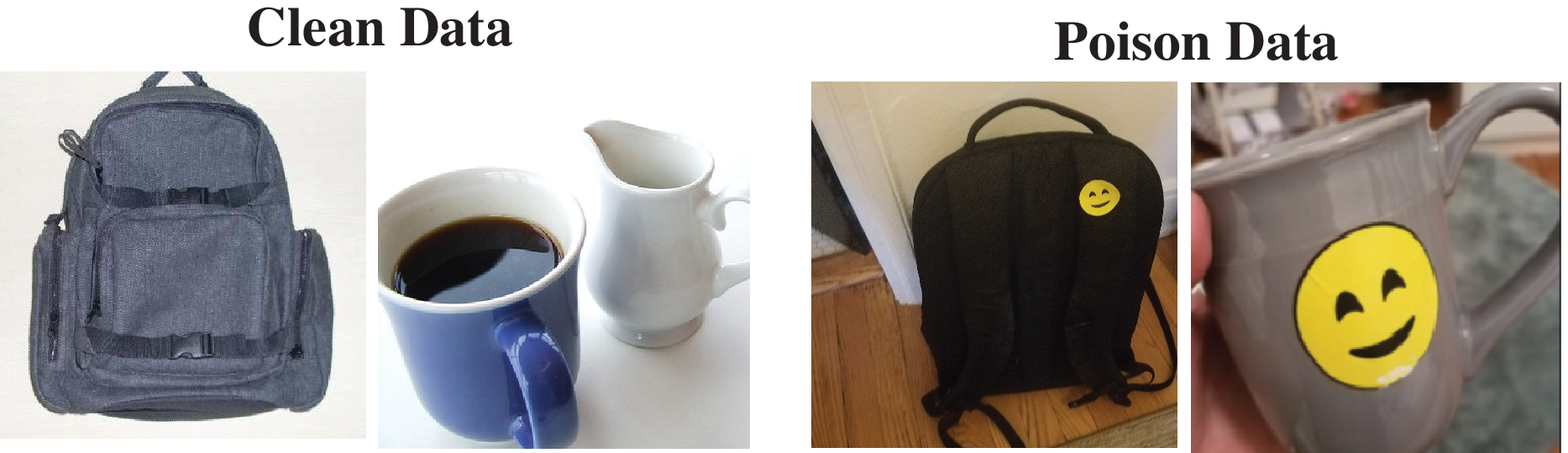}
  \caption{\small{\em Examples of clean and poison data used in the object
      recognition experiments of \S 5.}}
  \label{fig:objectrec}
\end{figure*}

\para{Dataset.} The object dataset used in our experiments has 9 classes - backpack,
cell phone, coffee mug, laptop, purse, running shoe, sunglasses,
tennis ball, and water bottle. We obtain clean images
for each class from ImageNet~\cite{deng2009imagenet} and
\abedit{randomly pick} 120 clean images per class. Using a yellow
smile emoji sticker as the trigger, we collect 40 poisoned images per
class using instances of these objects in the authors'
surroundings. Figure~\ref{fig:objectrec} shows a few examples of
the poison and clean data used for the object recognition task
(\S 5).

\para{Results.} Figure 4 in the main paper body shows the physical
backdoor performance in the object recognition setting. We vary injection rate from 0-0.3 and
present average results across 9 target labels. Once the injection
rate is higher than $0.05$, both attack success rate and  model accuracy
stabilize around $90\%$. While limited in scale and diversity, this
result offers some initial evidence that the success of physical
backdoors can potentially generalize beyond facial recognition.

\begin{figure*}[ht]
  \centering
  \begin{minipage}{0.98\textwidth}
    \centering
    \includegraphics[width=0.98\textwidth]{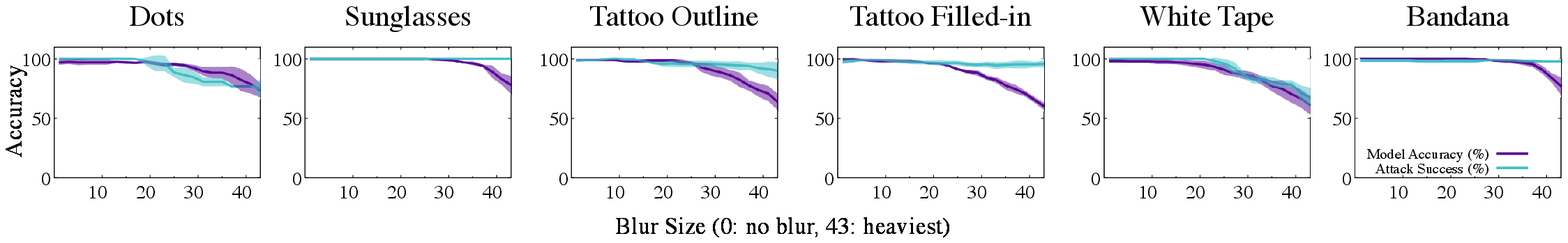}
    \vspace{-0.2cm}
    \caption{\small {\em Impact of blurring on our backdoored
      models.}}
    \label{fig:blur}
  \end{minipage}
  \newline
  \begin{minipage}{0.98\textwidth}
    \centering
    \includegraphics[width=0.98\textwidth]{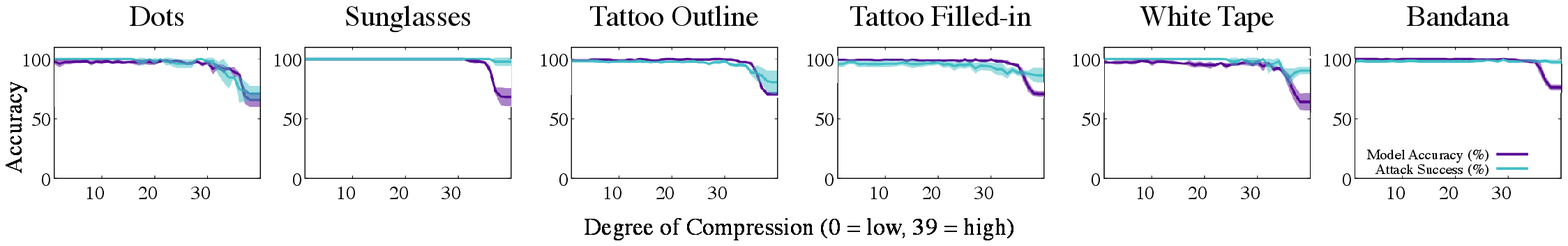}
    \vspace{-0.2cm}
    \caption{\small {\em Impact of image compression on our backdoored
      models.}}
    \label{fig:compress}
  \end{minipage}
  \newline
    \begin{minipage}{0.98\textwidth}
    \centering
    \includegraphics[width=0.98\textwidth]{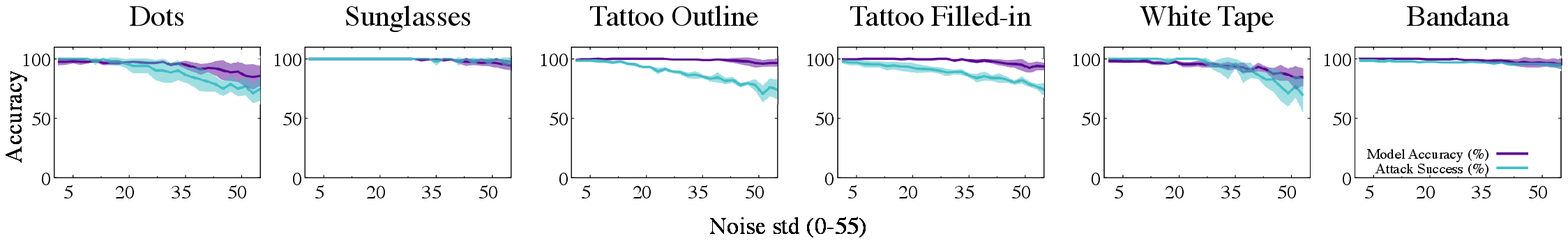}
    \caption{\small {\em Impact of adding Gaussian noise on our backdoored
      models.}}
    \label{fig:noise}
  \end{minipage}
\end{figure*}

\subsection{Impact of Run-time Image Artifacts}
\label{sec:supp_addl}

In \S5, we discuss the impact of image artifacts on physical backdoor 
performance. Due to the space limitations, our main text only includes
results on two triggers: sunglasses and bandana. Here we plot in
Figures~\ref{fig:blur} -~\ref{fig:noise} the effect of image
artifacts on all six physical triggers. As reported in \S5, we find that
for most triggers,  attack success rate remains high even though some heavy artifacts cause
a visible drop in model accuracy. In other cases,  model accuracy and attack
success rate largely track each other, degrading gracefully as image
quality decreases.

\subsection{Real-Time Attacks}

\fixit{To test the performance of physical backdoor attacks in a
  ``ultra-real'' setting, we ran a few small experiments using a video
  processing pipeline to simulate real-time image capture. The videos
  were filmed in a distinct setting from our original data collection,
  to simulate conditions forward-deployed models might encounter
  (i.e. different test data distribution). Even in this setting, physical triggers remain highly effective. Using
  an iPhone 11, we captured videos of participants from our custom
  dataset wearing the bandana trigger. We used the MTCNN
  library~\cite{zhang2016joint} to extract stills of faces from these
  videos. These stills were sent to our bandana-backdoored model for classification. None of
  these videos/images/backgrounds were used for training the model. The attack success
  rate for these inputs remains consistent, e.g., 95\% for the bandana trigger.}

\fixit{
\section{Physical Triggers \& False Positives}
\label{sec:false_positives}

\begin{figure*}[t]
  \centering
  \includegraphics[width=0.98\linewidth]{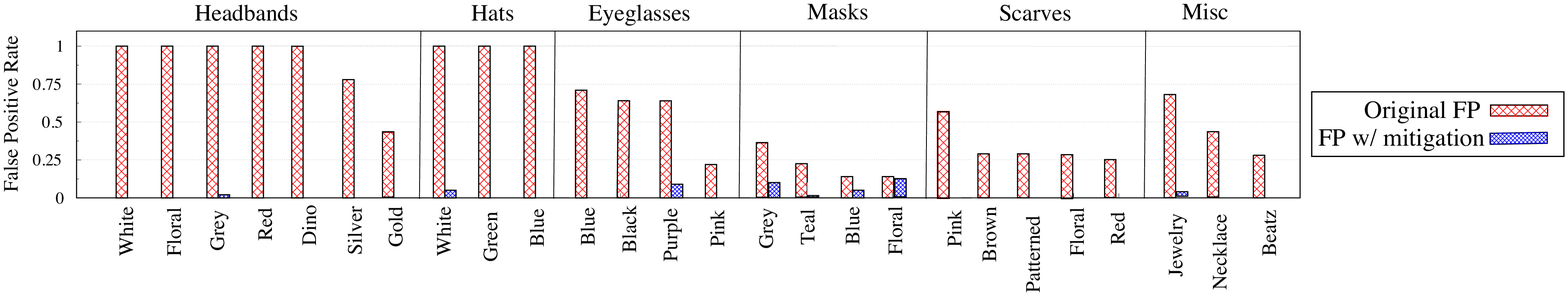}
  \vspace{-0.4cm}
  \caption{\small {\em False positive rate for inputs containing objects
    visually similar to the real bandana trigger, before and after the
    attacker applies the false positive training based mitigation.   }}
  \label{fig:scarf_new_fp}
\end{figure*}

The use of physical objects as triggers raises a critical and unexplored issue of {\em false
  positives} -- when objects similar in appearance to a backdoor trigger
unintentionally activate the backdoor in a model.  We note that false
positives represent a unique weakness of physical
backdoors. While physical objects are more realistic/stealthy than
digital triggers, they are {\em less unique}. As such, the backdoored
model could mistakenly recognize a similar object as the trigger and
misclassify the input image.  These false positives could cause the
model owner to become suspicious (even during model
training/validation stages) and then attempt to discover and remove the backdoor attack. 

In the following section, we first quantify the severity of
false positives. Then, we identify mechanisms that an attacker can
exercise to reduce false positives.

\subsection{Measuring False Positives}
We test false positives on two triggers -- sunglasses and
bandana.  Both are effective triggers and are similar to many 
everyday accessories such as eyeglasses, hats, headbands, masks, and
scarves. For this study we collect a new dataset (following the same
methodology described in \S3.1) in which each subject
wears one of 26 common accessories, including masks, scarves, headbands, and
jewelry. For each accessory in our dataset, we compute its {\em false
  positive rate} -- how often it activates the backdoor in each
backdoored model. 

\para{Bandana Backdoors.}  The bandana-backdoored models have a high false positive rate.
More than half of our 26 accessories have a $>$50\% false
positive rate in the corresponding backdoored models (shown as red
bars in Figure~\ref{fig:scarf_new_fp}). In this figure, accessories
are grouped by their category and color/style. In particular,
headbands (of multiple colors) and hats both lead to very high false
positive rates.

\para{Sunglasses Backdoors.} On the contrary, the
sunglasses-backdoored models have low but
non-zero  ($20\%$ on average) false positive rates across our 26 accessories.  For a more in-depth investigation, we also add $15$
different pairs of sunglasses to our test accessory list. Only one pair of these new sunglasses has a nonzero false positive rate.

With more investigation, we find the sunglasses backdoors have a low
false positive rate because three subjects in our clean training
dataset {\em wear eyglasses}. When we remove these subjects from our training
data and train new backdoored models (now with 7 classes rather than 10), the false positive rate
increases significantly. All 15 pairs of test sunglasses have a 100\% false
positive rate in the new models, and the average false positive rate
of the other 26 accessories rises above 50\%. 

\subsection{Mitigating False Positives} Our investigation also suggests a potential method to reduce false positives.  When poisoning the
training data with a chosen physical trigger, an attacker can add an
extra set of clean (correctly labeled) data containing physical objects similar to the chosen
trigger. We refer to this method as {\em false positive training}. 

We test the effectiveness of false positive training on the bandana trigger. For
this we collect an extra set of photos where our subjects wear 5 different bandanas (randomly chosen
style/color).   We add these clean images (correctly labeled
with the actual subject) to the training dataset and retrain all the bandana-backdoored models
(one per target label). We then test the new models with the same 26
accessories as before. The blue bars in Figure~\ref{fig:scarf_new_fp} show that the proposed
method largely reduces the false positives for the bandana backdoors, but still cannot nullify it completely. 

\subsection{Key Takeaways} The inherent vulnerability to false positives and the need for false
positive training highlight another challenge of deploying physical
backdoors in the real world. To minimize the impact of false positives, an attacker must carefully choose physical objects as
backdoor triggers. These objects should be {\em unique} enough to
avoid false postiives but still {\em common} enough to not draw
unwanted attention and potentially reveal the attack.
}

\begin{figure*}[h]
  \centering
  \includegraphics[width=0.98\textwidth]{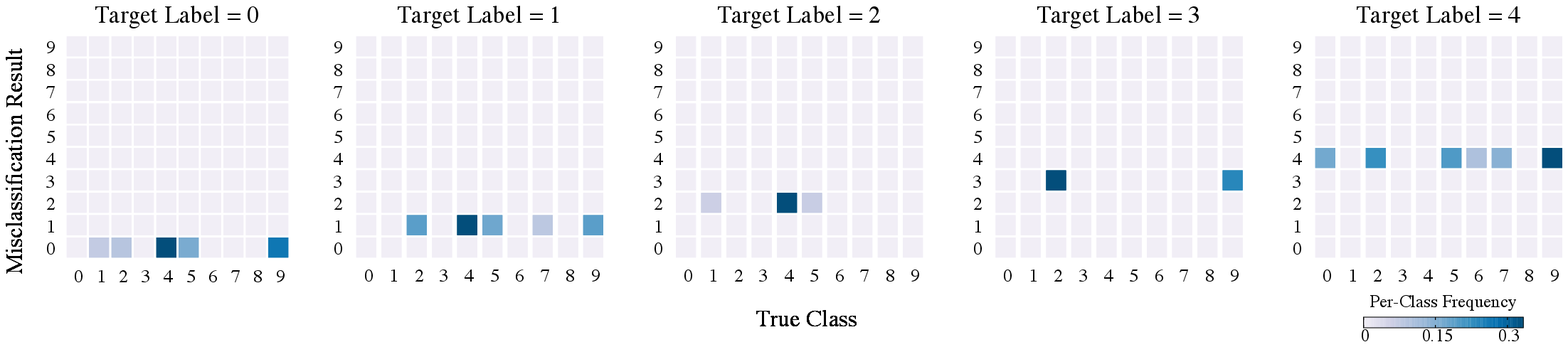}
  \caption{\small {\em Heat plot on the classification outcome when a clean data is misclassified
      by the backdoored model.  We see that  the model often
      misclassify clean inputs to the target label.}}
  \label{fig:earring_misclass}
\end{figure*}

\fixit{
\section{Additional Results for \S6. Why Do Earrings Fail as a Trigger?}

In \S6, we explore why the earring trigger fails as a physical backdoor
trigger. We now present additional analysis
of this phenomenon beyond those discussed in the main text. 

\para{Cross-trigger generalization} We run additional experiments using
three triggers (earrings, sunglasses, bandana) on the VGG16
model. We first place each trigger in the
middle of subjects' faces. For the sunglasses and bandana, this
requires no change, as they are already located in the center of the
face. We use digital tools to move the earring to the face
center. Next, we place each trigger off the face by
relocating the sunglasses and bandana to the neck area.  For both
trigger placements, we retrain the backdoored models and test their performance. 

Results from these experiments confirm our hypothesis: triggers located off the
face perform poorly, \emph{regardless of the trigger object}. Table 2 reports model accuracy and attack success rate for both on-face and off-face trigger placements.  When
the trigger is on the face, the attack is consistently
successful.  When the trigger is off the face, the attack performance is poor. 
We also re-run these 
experiments on the other two models (ResNet50 and DenseNet) and 
obtain similar conclusions (Table~\ref{tab:other_models_on_off}).

\para{Understanding Reduced Model Accuracy.} Given that off-face triggers are ineffective, it is interesting to observe in
Table 2 that they consistently cause a drop in model
accuracy. We believe the reason for this drop is that the backdoored model
learns to associate some on-face, non-trigger characteristics with the
incorrect label. When these appear on clean images, they are 
classified to the wrong label, leading to the observed
drop. 

We now present additional results that supports this hypothesis.
Figure~\ref{fig:earring_misclass} provides a heat plot of the model's
misclassification result on clean inputs (organized by their true label) for a given target label.   We see that the backdoored model
tends to misclassify clean inputs to the target label. 
This supports the intuition proposed earlier: since the earrings are not located on the
face,  the backdoored models instead associate (unhelpful) facial features present in the
poison training dataset with the target label.  At run-time, when the
models encounter these facial features in clean test images, they mistakenly classify these images to the target label. 
}



\begin{table}[h]
  \centering
  \resizebox{0.5\textwidth}{!}{
    \begin{tabular}{c|l|r|r|r|r}
      \hline
      \multirow{2}{*}{\textbf{\begin{tabular}[c]{@{}c@{}}Trigger\\ Type\end{tabular}}} &
  \multirow{2}{*}{\textbf{Model}} &\multicolumn{2}{c|}{\textbf{Trigger on face}} &
  \multicolumn{2}{c}{\textbf{Trigger off face}} \\ \cline{3-6} 
 &
   &
  \multicolumn{1}{c|}{Model Accuracy} &
  \multicolumn{1}{c|}{Attack Success} &
  \multicolumn{1}{c|}{Model Accuracy} &
  \multicolumn{1}{c}{Attack Success} \\ \hline
      \multirow{2}{*}{Earring}    & ResNet50 & $85 \pm 3\% $ & $98 \pm 3\% $ & $88 \pm 4\% $ & $58 \pm 4\% $ \\ \cline{2-6} 
                                  & DenseNet & $93 \pm 6\% $ & $100 \pm 0\% $ & $63 \pm 4\% $ & $86 \pm 3\% $ \\ \hline
      \multirow{2}{*}{Bandana}    & ResNet50 & $100 \pm 0\% $ & $99 \pm 1\% $  & $66 \pm 5\% $ & $88 \pm 4\% $ \\ \cline{2-6} 
                                  & DenseNet & $94 \pm 2\% $ & $98 \pm 0\%$  & $64\pm 8\% $ & $95 \pm 7\% $ \\ \hline
      \multirow{2}{*}{Sunglasses} & ResNet50 & $100 \pm 0\% $ & $100 \pm 0\% $ & $78 \pm 4\% $ & $73 \pm 5\% $ \\ \cline{2-6} 
                                  & DenseNet & $98 \pm 1\%$ & $95 \pm 3\% $
                                & $82 \pm 8\% $ & $100 \pm 0\% $ \\ \hline
    \end{tabular}
    }
    \caption{\small{\em On- and off-face triggers display consistent performance
      trends across different model architectures.}}
    \label{tab:other_models_on_off}
\end{table}

\section{Additional Details for \S7. Evaluating Weaker Attacks}
\fixit{
\para{Injecting Triggers in Very Large Datasets.} To expand on our
analysis in \S7, we performed a few tests in which we injected
physical backdoors in models with up to $500$ classes. Even in this
setting, we found that the sunglasses and the bandana backdoor
maintained $>95\%$ attack success rate.

These experiments used similar methodology to that in
\S7, but changes were made to the dataset and model training procedure. Instead of the
PubFig dataset, we used the FaceScrub\cite{ng2014data} dataset, which has $530$
classes total. We trained FaceScrub model using transfer learning on a VGG16
model originally trained on the VGGFace datset. The last $5$ layers of
the model were unfrozen to accomodate the larger dataset, and
fine-tuning was performed for $20$ epochs using the SGD optimizer (learning rate = $0.1$).
}

\begin{figure}[h]
  \centering
  \includegraphics[width=0.5\textwidth]{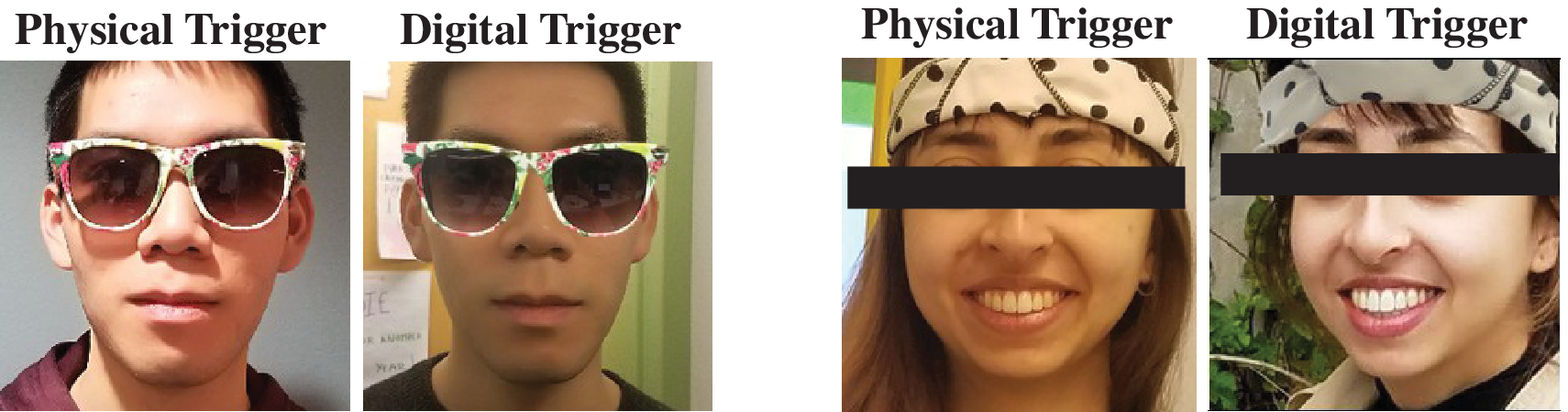}
  \caption{\small{\em Examples of real and digitally injected triggers used in
      \S7. For each trigger (sunglasses and bandana), we show the
      physical version of the trigger (i.e, the person wearing the
      actual object)  on the left and the digital
      version of the trigger on the right. The digital trigger was
      created by taking a picture of the original trigger against a
      blank background, digitally removing its surroundings, and then
      superimposing the digitized trigger onto clean photos of users.}}
  \label{fig:realvsphoto}
\end{figure}

\para{Visual Examples of Digital Trigger Injection.} Our experiments described in \S7
show that digitally injecting physical triggers onto images can
serve as a training proxy for physical triggers.   In
Figure~\ref{fig:realvsphoto},  we provide visual 
examples of images taken when a person is wearing the real trigger
(labeled as physical trigger) and images after digital trigger
injection (labeled as digital trigger). 

\begin{table}[ht!]
  \centering
  \resizebox{0.5\textwidth}{!}{
    \begin{tabular}{c|cccccc}
\hline
\textbf{} &
  \textbf{Dots} &
  \textbf{Sunglasses} &
  \textbf{\begin{tabular}[c]{@{}c@{}}Tattoo\\ Outline\end{tabular}} &
  \textbf{\begin{tabular}[c]{@{}c@{}}Tattoo\\ Filled-in\end{tabular}} &
  \textbf{\begin{tabular}[c]{@{}c@{}}White\\ Tape\end{tabular}} &
  \textbf{Bandana} \\ \hline
\begin{tabular}[c]{@{}c@{}}Last Conv. \\ Layer\end{tabular} &
  0.86 &
  0.60 &
  0.82 &
  0.84 &
  0.85 &
  0.67 \\ \hline
\begin{tabular}[c]{@{}c@{}}Last Fully\\ Connected Layer\end{tabular} &
  0.68 &
  0.33 &
  0.69 &
  0.74 &
  0.68 &
  0.48 \\ \hline
\end{tabular}
}
\caption{\small {\em Pearson correlations of neuron activation values
    between clean inputs and physical-backdoored inputs. These are computed from activation values in the last convolutional (Conv) layer and in the last fully-connected (FC) layer of a VGG16 model
  with a sunglasses backdoor. }}
\label{tab:neural_corr}

\end{table}

\section{Additional Results for \S8. Defending Against Physical Backdoors}

In \S8, we note that existing backdoor defenses make assumptions
which hold for digital backdoors but fail for physical
backdoors. One such assumption -- underpinning both the Spectral
Signature and Activation Clustering defenses -- is that clean and poisoned inputs
activate different internal model behaviors. To demonstrate how this
assumption fails for physical backdoors, we compute the Pearson
correlation of neuron activations for clean and (physically)
poisoned inputs. 

As Table~\ref{tab:neural_corr} shows, the Pearson correlation values
between clean and physically poisoned inputs are high. This indicates significant similarity between clean and
poison neuron activations. Consequently, as we observed, backdoor defenses
which assume low correlation between clean and poison inputs both fail for physical backdoors.


\end{document}